%% file: Template.tex
\documentclass{article}
\usepackage{spconf,amsmath,graphicx}

\usepackage{booktabs}
\usepackage{makecell}
\usepackage{array}
\usepackage{multicol}
\usepackage{multirow}
\usepackage{amssymb}
\usepackage{pifont}
\usepackage{tabularx}
\usepackage{inputenc}
\usepackage{xcolor}
\usepackage{amsmath}
\usepackage{bm}
\usepackage{colortbl}
\usepackage{booktabs}
\usepackage{url}
\usepackage{amsmath}

\title{HVD: Human Vision-Driven Video Representation Learning \\ for Text-Video Retrieval}
\name{Zequn Xie$^{1}$$^\star$ \thanks{$^\star$ Equal contribution.} \qquad Xin Liu$^{2}$$^\star$ \qquad Boyun Zhang$^{}$\qquad Yuxiao  Lin$^{1}$ \qquad Sihang Cai$^{1}$ \qquad Tao Jin$^{1}$$^\ddagger$ \thanks{$^\ddagger$ Corresponding author.}}
\address{
	$^{1}$ Zhejiang University \\
	$^{2}$ Southwestern University of Finance and Economics}

\begin{document}
\ninept
\maketitle

\begin{abstract}
The success of CLIP has driven substantial progress in text-video retrieval. However, current methods often suffer from ``blind'' feature interaction, where the model struggles to discern key visual information from background noise due to the sparsity of textual queries. To bridge this gap, we draw inspiration from human cognitive behavior and propose the \textbf{H}uman \textbf{V}ision-\textbf{D}riven (\textbf{HVD}) model. Our framework establishes a coarse-to-fine alignment mechanism comprising two key components: the \textbf{F}rame \textbf{F}eatures \textbf{S}election \textbf{M}odule (\textbf{FFSM}) and the \textbf{P}atch \textbf{F}eatures \textbf{C}ompression \textbf{M}odule (\textbf{PFCM}). FFSM mimics the human macro-perception ability by selecting key frames to eliminate temporal redundancy. Subsequently, PFCM simulates micro-perception by aggregating patch features into salient visual entities through an advanced attention mechanism, enabling precise entity-level matching. Extensive experiments on five benchmarks demonstrate that HVD not only captures human-like visual focus but also achieves state-of-the-art performance.
\end{abstract}

\input{introduce}

\section{Methods}

\begin{figure*}[ht!]
	\centering
	% left bottom right top
	\includegraphics[width=\linewidth, trim=30pt 30pt 30pt 240pt, clip]{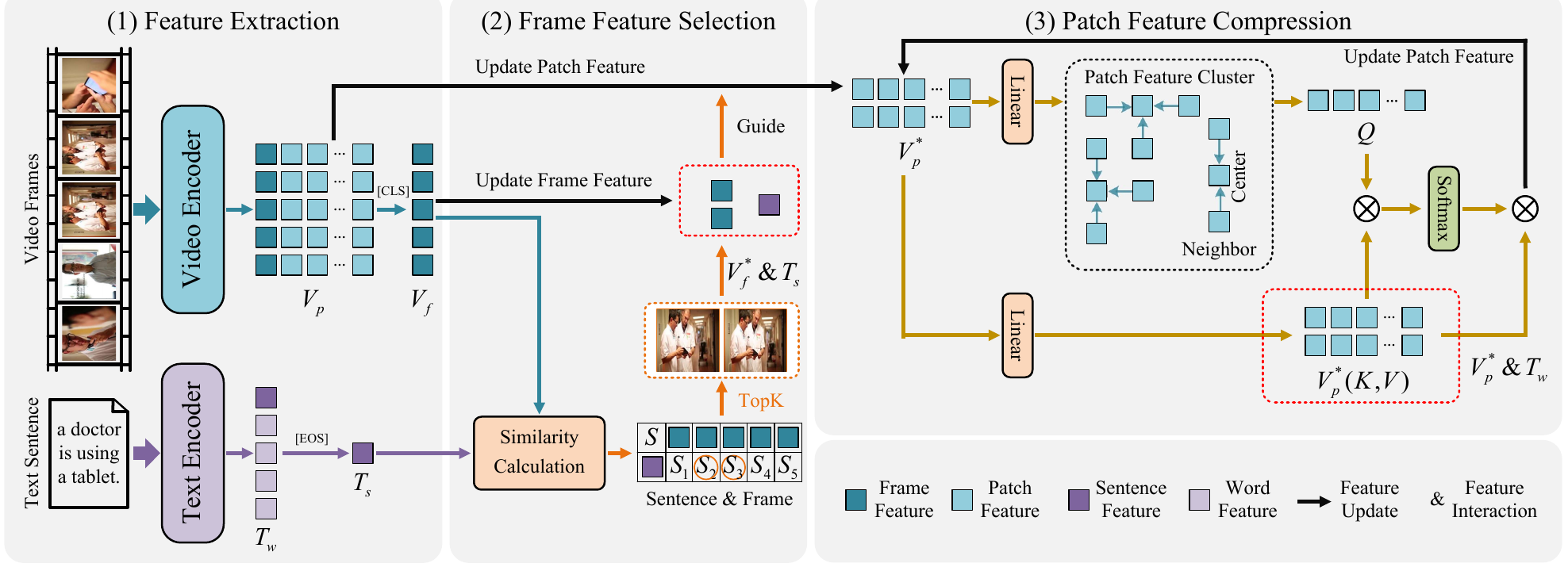}
	\caption{\textbf{Framework.} (1) CLIP Feature Extraction Module. (2) Frame Feature Selection Module: selects keyframes that best match the text from a human global perspective based on text-frame similarity. (3) Patch Feature Compression Module: compresses visual features from a human local perspective, focusing on the most informative patch regions.}
	\label{Fig-2}
\end{figure*}

\subsection{Preliminaries}

\noindent Let $\mathcal{D}=(\mathcal{T},\mathcal{V})$ denote a language and vision dataset, where $\mathcal{T}$ is a set of texts, and $\mathcal{V}$ is a set of videos. The goal of text-video retrieval is to rank the relevance between a text query $t \in \mathcal{T}$ and video set $\mathcal{V}$. Recent works \cite{Clip4clip,X-Pool,T-Mass} have shown CLIP's strong performance in modality feature representation, inspiring us to employ CLIP as our backbone. Specifically, a video $v \in \mathcal{V}$ consists of $N_f$ sequential frames $[f_1,f_2,...,f_{N_f}] \in {\mathbb{R}^{N_f \times H \times W \times C}}$, where each frame is divided into $N_p$ patches $[p_1,p_2,...,p_{N_p}] \in {\mathbb{R}^{N_p \times P \times P \times C}}$ with $P \times P$ size. Following CLIP \cite{CLIP}, we utilize the CLIP visual encoder to extract the patch features $V_p=[p_0,p_1,..., p_{N_p}] \in \mathbb{R}^{(N_p+1) \times D}$ of each frame, where $p_0$ represents the \texttt{[CLS]} token of the current frame. We aggregate the \texttt{[CLS]} tokens of all video frames to obtain the frame features $V_f=[f_1, f_2,..., f_{N_f}]\in \mathbb{R}^{N_f \times D}$. Similarly, given a text query $t \in \mathcal{T}$, we leverage the CLIP text encoder to extract word features $T_w=[w_\texttt{BOS}, w_1, w_2,..., w_{N_w}, w_\texttt{EOS}]\in \mathbb{R}^{(N_w+2) \times D}$, where $N_w$ denotes the length of word sequences, and \texttt{[BOS]} and \texttt{[EOS]} tokens mark the beginning and end of a text sequence, respectively. We take the representation of the \texttt{[EOS]} token as the sentence feature $T_s=[s] \in \mathbb{R}^{1 \times D}$. The overall video and text features extraction process is illustrated in Fig. \ref{Fig-2}(1).

Feature interaction refers to the process of computing the similarity between textual and visual features, and mean pooling is the most direct method, \textit{e.g.}, between $T_s$ and $V_f$:
\begin{equation}	
	S_{T_s,V_f} = \frac{\bm{T_s} \cdot \bm{V_f}}{||\bm{T_s}|| \cdot ||\bm{V_f}||},
	\label{Eq-1}
\end{equation}
where $\bm{V_f} = \frac{1}{N_f}\sum\nolimits_i^{N_f} {V_{f,i}}$. Therefore, the sentence-frame cross-modal contrastive loss $\mathcal{L}_{T_s,V_f}$ can be formulated as:
\begin{equation}
	\begin{aligned}
		\mathcal{L}_{T_s,V_f} &= -\frac{1}{2} \bigg( \frac{1}{B} \sum_{i=1}^{B} \log \frac{\exp(S_{T_s^i,V_f^i}/\tau)}{\sum_{j=1}^{B} \exp(S_{T_s^i,V_f^j}/\tau)} \\
		&\hspace{2.1em} + \frac{1}{B} \sum_{i=1}^{B} \log \frac{\exp(S_{T_s^i,V_f^i}/\tau)}{\sum_{j=1}^{B} \exp(S_{T_s^j,V_f^i} / \tau)} \bigg),
	\end{aligned}
	\label{Eq-2}
\end{equation}
where $B$ is the batch size, $\tau$ is the temperature hyper-parameter, and $S_{T_s^i,V_f^j}$ represents the similarity between the $i^{th}$ text sentence feature and the $j^{th}$ video frame feature in the entire batch. This loss function maximizes the similarity of positive pairs and minimizes the similarity of negative pairs. Similarly, $\mathcal{L}_{T_w,V_f}$, $\mathcal{L}_{T_s,V_p}$, and $\mathcal{L}_{T_s,V_f}$ can be computed.

\subsection{Frame Features Selection Module}

\noindent\hspace{1.20em} In this part, we propose a \textbf{F}rame \textbf{F}eatures \textbf{S}election \textbf{M}odule (\textbf{FFSM}) from a human macro alignment perspective, as shown in Fig. \ref{Fig-2}(2). First, when aligning text and video, we, as humans, first read the overall sentence feature $T_s=[s] \in \mathbb{R}^{1 \times D}$, and then compare it with each video frame features $V_f=[f_1,f_2,...,f_{N_f}] \in \mathbb{R}^{N_f \times D}$ to identify the effective frames $V_f^* \in \mathbb{R}^{N_f^* \times D}$, where $N_f^*<N_f$. Next, from a human-centric perspective, the selection of effective frames $V_f^*$ can be directly based on the similarity $S_{T_s,V_f}=\frac{{{s} \cdot {f}}}{{||{s}|| \cdot ||{f}||}} \in \mathbb{R}^{1 \times N_f}$ between $T_s$ and $V_f$.
Finally, we select the $\textit{top-}N_f^*$ frames $V_f^*$ from $V_f$ based on $S_{T_s,V_f}$:
\begin{equation}
	V_f^* = \{f_i\},{i \in \arg\max_{i} S_{T_s,V_f}, i=1,2,...,N_f^*}.
	\label{Eq-3}
\end{equation}
In addition, the patch features are also updated from $V_p$ to $V_p^*$.

\subsection{Patch Features Compression Module}

\noindent\hspace{1.20em} In this part, we propose a \textbf{P}atch \textbf{F}eatures \textbf{C}ompression \textbf{M}odule (\textbf{PFCM}) from a human micro alignment perspective, as shown in Fig. \ref{Fig-2}(3). First, we focus on the patch features $V_p^*=N_f^* \times [p_1, p_2, \dots, p_{N_p}] \in \mathbb{R}^{N_f^* \times N_p \times D}$ under the FFSM. This step, benefiting from global features selection, not only focuses on key frames but also significantly reduces the number of patch features. Next, we further compress the patch features $V_p^*$ from a human cognitive perspective.

Specifically, we utilize a variant of the $k$-nearest neighbor-based density peaks clustering algorithm (DPC-KNN) \cite{DPC-KNN}. Given patch features $V_p^*$, we compute the local density $\rho_i$ of each patch $p_i$ according to its $k$-neatest neighbors: $\rho _i = \exp ( - \frac{1}{k}\sum\limits {||{p_i} - {p_j}|} |_2)$, where ${{p_j} \in \text{KNN}({p_i})}$, and $\text{KNN}(p_i)$ denotes the $k$-neatest neighbors of the patch $p_i$. Then, we compute the distance indicator $\delta_i$ of each patch:

\begin{equation}
	\delta_i = 
	\begin{cases}
		\mathop {\min } ||{p_i} - {p_j}|{|_2},  & \rho_i < \rho_j, \\ 
		\mathop {\max } ||{p_i} - {p_j}|{|_2},  & \rho_i \ge \rho_j.
	\end{cases}
	\label{Eq-4}
\end{equation} 
Intuitively, the patch $p_i$ with a larger local density $\rho_i$ and distance indicator $\delta_i$ is more likely to become a cluster center. We determine a cluster center by selecting the patches with the highest scores $\rho_i \times \delta_i$, and then merge the neighboring patches. The merged patch $p_i^*$ is fed into a transformer block as query $Q$, and the original patch $p_i$ is used as key $K$ and value $V$ is added to the attention weight as follows:
\begin{equation} 
	\text{Attention}(Q,K,V)=\text{softmax}\left(\frac{QK^T}{\sqrt{D}}\right)V,
	\label{Eq-5}
\end{equation} 
By introducing the clustering algorithm and the attention mechanism, PFCM not only reduces the number of patch features but also focuses on key features and spatial relationships. Finally, we iteratively apply PFCM to compress and aggregate patch features, reducing their number.

\subsection{Training and Sampling}

\noindent\hspace{1.20em} To effectively integrate global and local visual information, we combine the macro-level frame features $V_f^*$ derived from FFSM and the micro-level patch features $V_p^*$ obtained from PFCM. Based on these representations, we compute the corresponding similarity scores: the sentence-frame similarity $S_{T_s,V_f^*}$ and the word-entity similarity $S_{T_w,V_p^*}$. The overall training objective is to jointly minimize the losses from both perspectives. Thus, we compute the total loss $\mathcal{L}_{total}$ as defined in Eq. \ref{Eq-6}:
\begin{equation}
	\mathcal{L}_{total} = \mathcal{L}_{T_s,V_f^*} + \mathcal{L}_{T_w,V_p^*}.
	\label{Eq-6}
\end{equation}
During the inference phase, we calculate the final similarity between the text query and the video by aggregating both the global similarity $S_{T_s,V_f^*}$ and the local similarity $S_{T_w,V_p^*}$ to determine the retrieval ranking.

\begin{figure*}[!h]
	\centering
	% left bottom right top
	\includegraphics[width=\linewidth, trim=20pt 20pt 20pt 20pt, clip]{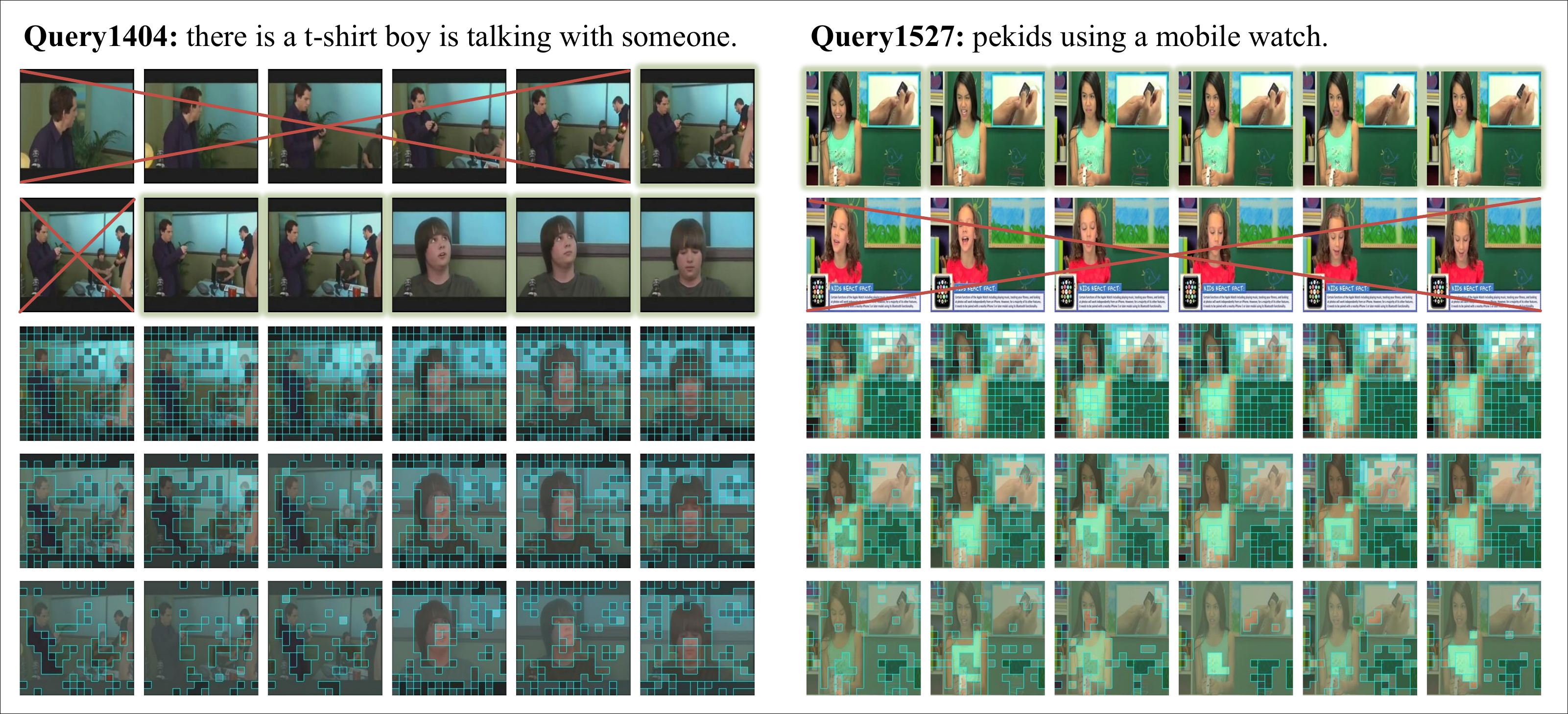}
	\caption{\textbf{Visualization}. \textcolor{red}{Red} lines indicate the frame features that are selected for removal.}
	\label{Fig-3}
\end{figure*}

\section{Experiments}

\subsection{Experimental Settings}

\noindent\hspace{1.20em} We evaluate HVD on five benchmarks: MSRVTT, DiDeMo, LSMDC, ActivityNet, and Charades. Performance is reported using R@K (K=1, 5, 10), MdR, and MnR. Utilizing CLIP \cite{CLIP} as the backbone, we train the model for 5 epochs with a batch size of 32 and a feature dimension of 512. The input is configured with $N_f=12$ sampled frames and a maximum text length of $N_w=24$. Both frame and patch retention ratios are set to 0.5. Additional implementation details follow standard protocols \cite{Clip4clip,X-Pool,HBI,T-Mass,EERCF}.

\subsection{Experimental Performance}

\noindent\hspace{1.20em} In Tab. \ref{Tab-1}, we report the retrieval performance of HVD on the MSRVTT dataset, achieving an R@1 of 48.8. Compared with HBI \cite{HBI}, this represents an improvement of 0.2, indicating that our method benefits from visual feature selection and compression. In Tab. \ref{Tab-2}, we further evaluate the retrieval performance of HVD on datasets such as DiDeMo. HVD achieves consistent performance improvements on both the long-video DiDeMo and the short-text LSMDC.

\begin{table}[!ht]
	\centering
		\begin{tabular}{l|@{\hskip 5pt}c@{\hskip 5pt}c@{\hskip 5pt}c@{\hskip 5pt}c@{\hskip 5pt}c}
		\toprule
		\multirow{2.7}{*}{{Methods}} & 
		\multicolumn{4}{c}{{MSRVTT (Text$\to$Video)}} \\
		\cmidrule(lr){2-6}   
		& {R@1}$\uparrow$ & {R@5}$\uparrow$ & {R@10}$\uparrow$ & MdR$\downarrow$ & {MnR}$\downarrow$ \\
		\midrule
		Clip4clip \cite{Clip4clip} & 44.5 & 71.4 & 81.6 & 2.0 & 15.3 \\
		X-Pool \cite{X-Pool} & 46.9 & 72.8 & 82.2 & 2.0 & 14.3 \\
		UATVR \cite{UATVR} & 47.5 & 73.9 & 83.5 & 2.0 & 12.3 \\
		HBI \cite{HBI} & 48.6 & 74.6 & 83.4 & 2.0 & 12.0 \\
		EERCF \cite{EERCF} & 47.8 & 74.1 & 84.1 & 2.0 & - \\
		BiHSSP \cite{BiHSSP} & 48.1 & 74.0 & 84.1 & 2.0 & 12.1 \\
		\rowcolor{red!5}
		{HVD (Ours)}  & \textbf{48.8} & \textbf{75.2} & \textbf{85.3} & \textbf{2.0} & \textbf{11.7} \\
		\bottomrule
	\end{tabular}
	\caption{Retrieval performance on the MSRVTT datasets. ``$\uparrow$'' means that higher is better. ``$\downarrow$'' means that lower is better.}
	\label{Tab-1}
\end{table}

\begin{table}[ht!]
	\centering
	\begin{tabular}{l|@{\hskip 7pt}c@{\hskip 7pt}c|@{\hskip 7pt}c@{\hskip 7pt}c}
		\toprule
		\multirow{2.7}{*}{Methods} & \multicolumn{2}{c|}{DiDeMo} & \multicolumn{2}{c}{LSMDC} \\
		\cmidrule(lr){2-3} \cmidrule(lr){4-5}
		& R@1$\uparrow$ & R@10$\uparrow$ & R@1$\uparrow$ & R@10$\uparrow$ \\
		\midrule
		EMCL-Net \cite{EM} & 45.3 & 82.3 & 23.9 & 53.7 \\
		X-Pool \cite{X-Pool} & 44.6 & 82.0 & 25.2 & 53.5 \\
		CLIP-VIP \cite{CLIP-VIP} & 48.6 & 84.4 & 25.6 & 54.4 \\
		DiCoSA \cite{DiCoSA} & 45.7 & 83.5 & 25.4 & 54.0 \\
		\rowcolor{red!5}
		HVD (Ours) & \textbf{48.9} & \textbf{84.0} & \textbf{26.1} & \textbf{57.0} \\
		\midrule
		\multirow{2.7}{*}{Methods} & \multicolumn{2}{c|}{ActivityNet} & \multicolumn{2}{c}{Charades} \\
		\cmidrule(lr){2-3} \cmidrule(lr){4-5}
		& R@1$\uparrow$ & R@10$\uparrow$ & R@1$\uparrow$ & R@10$\uparrow$ \\
		\midrule
		Clip4clip \cite{Clip4clip} & 40.5 & 83.6 & 9.9 & 36.8 \\
		HBI \cite{HBI} & 42.2 & 84.6 & - & - \\
		T-Mass \cite{T-Mass} & - & - & 14.2 & 48.3 \\
		\rowcolor{red!5}
		HVD (Ours) & \textbf{44.6} & \textbf{85.7} & \textbf{18.8} & \textbf{54.6} \\
		\bottomrule
	\end{tabular}
	\caption{Retrieval performance on other datasets.}
	\label{Tab-2}
\end{table}

In Tab. \ref{Tab-3}, we perform an ablation study on HVD's FFSM and PFCM, and using $\mathcal{L}_{T_s,V_f}$ as the baseline in Eq. \ref{Eq-1} (R1). The experiments show that combining macro frame selection (R2) with micro patch compression (R3) achieves the best retrieval performance (R4). This further indicates a close relationship between FFSM and PFCM: the former provides refined information to the latter, while the latter complements the former with additional details.

\begin{table}[ht!]
	\centering
	\begin{tabular}{c|@{\hskip 5pt}c@{\hskip 5pt}c|@{\hskip 5pt}c@{\hskip 5pt}c@{\hskip 5pt}c@{\hskip 5pt}c}
		\toprule
		Rx & FFSM & PFCM & {R@1}$\uparrow$ & {R@5}$\uparrow$ & {R@10}$\uparrow$ & {MnR}$\downarrow$ \\
		\midrule
		R1 & & & 44.6 & 71.0 & 81.7 & 15.3 \\
		\midrule
		R2 & \ding{51} & & 44.7 & 72.5 & 82.4 & 14.6 \\
		R3 & & \ding{51} & 46.3 & 74.8 & 83.7  & 12.2 \\
		\rowcolor{red!5}
		R4 & \ding{51} & \ding{51} & \textbf{48.8} & \textbf{75.2} & \textbf{85.3} & \textbf{11.7} \\
		\bottomrule	
	\end{tabular}
	\caption{Ablation studies of FFSM and PFCM.}
	\label{Tab-3}
\end{table}

\begin{table}[ht!]
	\centering
	\begin{tabular}{c|cccc|cc}
		\toprule
		\multirow{2.7}{*}{{($\hbar$, $\mathchar'26\mkern-10mu\lambda$)}} & 
		\multicolumn{4}{c|}{$N_f^*=\hbar N_f$,$N_p^*=\mathchar'26\mkern-10mu\lambda N_p$} & 
		\multicolumn{2}{c}{MSRVTT} \\  
		\cmidrule(lr){2-5} \cmidrule(lr){6-7} 
		& $N_f^*$ & $N_p^*$ & $N_p^*$ & $N_p^*$ & R@1$\uparrow$ & R@10$\uparrow$ \\
		\midrule
		(1.00, 1.00) & 12 & 600 & 600 & 600 & 44.6 & 81.7 \\
		\midrule
		(0.25, 0.50) & 3 & 90 & 45 & 23 & 44.2 & 81.5 \\
		(0.50, 0.75) & 6 & 225 & 169 & 127 & 46.2 & 84.7 \\
		\rowcolor{red!5}
		(0.50, 0.50) & 6 & 150 & 75 & 38 & \textbf{48.8} & \textbf{85.3} \\
		(0.50, 0.25) & 6 & 75 & 19 & 5 & 47.1 & 84.2 \\
		(0.75, 0.50) & 9 & 225 & 113 & 56 & 48.3 & 84.7 \\
		\rowcolor{red!5}
		\bottomrule
	\end{tabular}
	\caption{Ablation studies of factor $\hbar$ and factor $\mathchar'26\mkern-10mu\lambda$.}
	\label{Tab-4}
\end{table}

To further investigate the impact of feature selection and compression levels on retrieval performance, we set them to $\hbar$ and $\mathchar'26\mkern-10mu\lambda$, respectively, as shown in Tab. \ref{Tab-4}. $\hbar$ and $\mathchar'26\mkern-10mu\lambda$ denote the proportions of retained video frames and image patches, respectively. When setting ($\hbar, \mathchar'26\mkern-10mu\lambda$)= (0.50, 0.50), HVD achieves the best retrieval performance, further indicating that selecting an appropriate granularity can enhance feature interactions. Based on this parameter setting, we visualize the process of video frame selection and patch feature compression during training in Fig. \ref{Fig-3}. The non-red regions indicate frames used for text interaction. The last three rows show the results of three consecutive PFCM operations, with the number of features involved in the interaction gradually decreasing.

\section{Conclusion}

In this paper, we propose HVD, a human-vision-driven video representation model that integrates frame selection (FFSM) and patch compression (PFCM) to improve text–video retrieval performance. Guided by human macroscopic and microscopic perspectives, these modules jointly improve video representations at frame and patch levels, addressing both coarse- and fine-grained alignment. We evaluate HVD on five benchmark datasets, including MSRVTT, DiDeMo, LSMDC, ActivityNet, and Charades, achieving state-of-the-art retrieval performance. In addition, we further validate the effectiveness and intuitiveness of the proposed modules through ablation studies on FFSM, PFCM, and hyper-parameter settings, as well as visualizations. We hope that this work will provide inspiration to the video retrieval community.
     

% \clearpage
\bibliographystyle{IEEEbib}
\bibliography{strings,refs}

\end{document}

%% file: introduce.tex
\section{Introduction}

\noindent\hspace{1.20em} With the rise of online video platforms and the increasing focus on integrated planning and multimodal intelligent agents \cite{liu2025integratedplanningmachinelevelscheduling, he2025unified, kang2026multimodalmultiagentempoweredlegal, lu2025dammfnd, tong2025dapt, liu2026health, sun2025objective, zeng2025FSDrive, lu2024mace, lu2023tf, lu2024robust}, a large number of unlabeled videos are uploaded and downloaded. The emergence of text-video retrieval tasks has enabled these videos to be widely utilized and accessed. Recently, with the significant success of the large-scale text-image pretraining model CLIP \cite{CLIP} in representation learning, existing methods \cite{Clip4clip,X-Pool} typically adopt CLIP to project text and video into a shared latent space, thereby establishing feature-level similarity relationships. To achieve more precise feature matching, current methods \cite{HBI,UATVR,xie2026delvingdeeperhierarchicalvisual,zhang2024cf} improve retrieval performance by leveraging granularity alignment and feature enhancement techniques. However, are these methods sufficient in terms of granularity alignment accuracy and feature enhancement effectiveness?

\begin{figure}[!t]
	\centering
	% left bottom right top 4.0in
	\includegraphics[width=\linewidth, trim=15pt 15pt 15pt 15pt, clip]{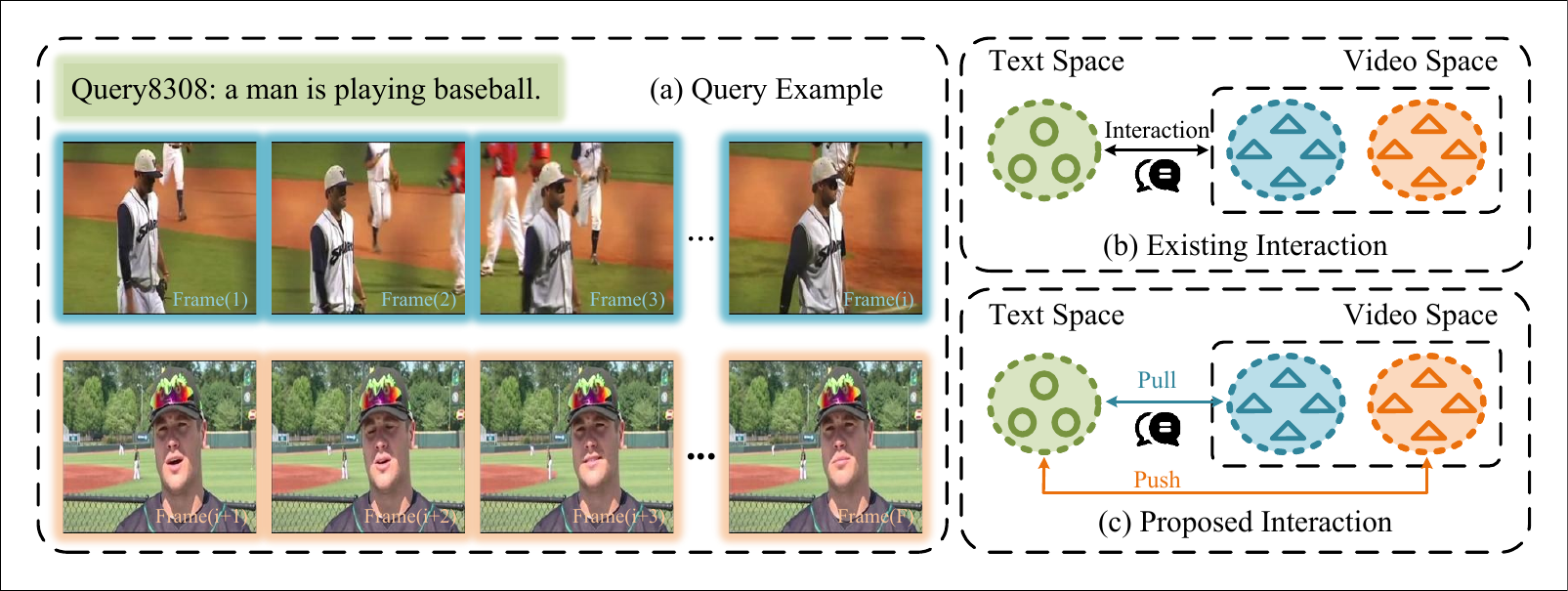}
	\caption{\textbf{Motivation.} Not all visual features are relevant to the text query. Blue borders indicate the relevant frames, while yellow borders indicate the irrelevant frames.}
	\label{Fig-1}
\end{figure}

As illustrated in Fig. \ref{Fig-1}(a), given the text query \textit{``a man is playing baseball''} and video frames, the frames show scenes of the man playing baseball as well as being interviewed. It is clear that only the first $i$ frames (\textcolor{blue}{blue borders}) are related to the text query, while the remaining frames (\textcolor{yellow}{yellow borders}) are unrelated. In the case of sparse queries, existing methods \cite{DiCoSA,HBI,xie2025chat} typically incorporate all visual features into the interaction with text without considering feature effectiveness in Fig. \ref{Fig-1}(b). From the perspective of repeated human comparison, we focus more on the scenes related to ``\textit{baseball}'' rather than the ``\textit{interview}.'' Meanwhile, after identifying the key frames, it is necessary to repeatedly compare the keywords in the query with the visual entities in the frames, such as ``\textit{man}'' or ``\textit{baseball}.'' In other words, through this iterative comparison process, we can determine the optimal cross-modal feature interaction in Fig. \ref{Fig-1}(c). Therefore, we summarize this process into two stages: \ding{182} from a macro perspective, we observe all the frames of the video as a whole, selecting key frames that are highly relevant to the text query and thereby eliminating unrelated scenes; \ding{183} from a micro perspective, based on the key frames, we gradually extract key visual entities from the video frames, repeatedly comparing them with the keywords in the text query to refine feature interactions and ensure the accuracy of cross-modal feature alignment.

Existing methods \cite{UCoFiA,T-Mass} have also conducted extensive explorations based on the above analysis. For example, UCoFiA \cite{UCoFiA} proposes a unified coarse-to-fine alignment model that effectively improves text-video retrieval performance from a comprehensive perspective. T-Mass \cite{T-Mass} adopts a similarity-based feature modeling approach to expand text features, rather than directly aggregating frame features. In addition, feature enhancement methods \cite{UATVR,Mv-adapter,EERCF} expand the receptive field. Although these methods achieve impressive retrieval performance, they fail to capture human interactive cognition and thus suffer from blind feature interaction. Specifically, \ding{182} Feature enhancement lacks robustness: when faced with distracting visual cues and sparse textual information, the enhanced features are affected by noise and lack stability. \ding{183} Feature interaction lacks precision: cross-modal feature matching and alignment are not accurate enough, resulting in blind spots or insufficient interaction of information.

To truly simulate human cognitive processes of alignment, we propose a novel text-video retrieval framework, named the \textbf{H}uman \textbf{V}ision-\textbf{D}riven (\textbf{HVD}) model. Fig. \ref{Fig-2} illustrates the overall framework. \textbf{First}, we propose a \textbf{F}rame \textbf{F}eatures \textbf{S}election \textbf{M}odule (\textbf{FFSM}) to accommodate the human macro-level alignment process. FFSM selects key video frames related to the text query based on similarity, effectively eliminating redundant and irrelevant frames while retaining the patch features necessary for fine-grained alignment, thereby achieving a certain level of feature accuracy at the macro level. \textbf{Second}, we propose a \textbf{P}atch \textbf{F}eatures \textbf{C}ompression \textbf{M}odule (\textbf{PFCM}) to align with the human micro-level alignment process. The core of PFCM lies in compressing the patch features from the valid frames selected by FFSM, and then extracting the visual entity features that align with the textual word features. Specifically, we use the K-Nearest-Neighbor-based density peaks clustering algorithm, \textit{i.e.}, DPC-KNN~\cite{DPC-KNN}, to aggregate the patch features. Meanwhile, to emphasize important patches and capture the spatial relationships between patches, we introduce importance scores and attention weights to re-represent the patch features. Finally, through repeated aggregation and re-representation, we can obtain entity features at the word feature level. \textbf{Third}, the two modules work collaboratively to achieve video feature selection and compression, eliminating the need to consider complex text alignment strategies in subsequent stages. Experiments on five benchmarks demonstrate that HVD achieves state-of-the-art retrieval performance.